\documentclass{article}

\usepackage{microtype}
\usepackage{graphicx}
\usepackage{subfigure}
\usepackage{booktabs}
\usepackage{amsmath}
\usepackage{amssymb}
\usepackage{placeins}

\usepackage{hyperref}
\usepackage{xurl}

\newcommand{\hr}[1]{\textcolor{black}{#1}}

\usepackage[preprint]{mlsys2025}

\mlsystitlerunning{HiLo-Token: Input-Adaptive High--Low Frequency Token Compression for Efficient Image Editing Model Serving}

\begin{document}

\twocolumn[
\mlsystitle{HiLo-Token: Input-Adaptive High--Low Frequency Token Compression for Efficient Image Editing Model Serving}

\mlsyssetsymbol{equal}{*}
\mlsyssetsymbol{adobework}{$\dagger$}

\begin{mlsysauthorlist}
\mlsysauthor{Haoran You}{adobe,purdue,adobework}
\mlsysauthor{Yotam Nitzan}{adobe}
\mlsysauthor{Lingzhi Zhang}{adobe}
\mlsysauthor{Yifan Gong}{adobe}
\mlsysauthor{Mang-Tik Chiu}{adobe}
\mlsysauthor{Connelly Barnes}{adobe}\\
\mlsysauthor{Yan Kang}{adobe}
\mlsysauthor{Yuqian Zhou}{adobe}
\mlsysauthor{Haitian Zheng}{adobe}
\mlsysauthor{Eli Shechtman}{adobe}
\mlsysauthor{Sohrab Amirghodsi}{adobe}
\end{mlsysauthorlist}

\mlsysaffiliation{adobe}{Adobe Inc.}
\mlsysaffiliation{purdue}{Purdue University}

\mlsyscorrespondingauthor{Haoran You}{haoranyou98@gmail.com}
\mlsyscorrespondingauthor{Sohrab Amirghodsi}{tamirgho@adobe.com}

\mlsyskeywords{Machine Learning, MLSys, Diffusion Transformers, Token Compression, Image Editing}

\vskip 0.3in

\begin{abstract}
Creative image editing tools, such as Adobe Photoshop's Remove and Generative Fill buttons, are central to everyday customer use and account for a major share of traffic. However, current generative AI models face significant latency challenges, which become even more pronounced as the field transitions from convolution-based U-Nets to Diffusion Transformers (DiTs). In our evaluation on hundreds of representative image editing samples spanning a wide range of mask ratios, the DiT module alone accounts for an average of 73\% of the total model latency, even after being distilled from 50 timesteps down to 8 timesteps.
To tackle this challenge, we propose \textbf{HiLo-Token}, an input-adaptive token compression framework that allocates a larger token budget to high-frequency, rich-context regions while assigning fewer tokens to low-frequency areas. Specifically, for the editing region specified by the user mask, we retain all tokens within a dilated mask to preserve strong locality and contextual relevance. Outside the editing region, we introduce a simple yet effective high-frequency token selection strategy based on spatial frequency to capture important local details, while using tokens from a heavily downsampled image to represent low-frequency components and preserve the coarse global structure.
Extensive experiments on production-level evaluation data validate the effectiveness of the proposed method, which achieves 3.13$\times$, 2.59$\times$, and 1.67$\times$ DiT speedups on A100-80GB for image editing tasks across small, medium, and large mask ratio categories with average ratios of 6.38\%, 15.92\%, and 35.36\%, respectively, without any regression in generation quality. In production, where the editing pipeline is served as disaggregated, independently autoscaled GPU workers, these per-request savings translate into a 33\% reduction in the GPUs required to serve the models.
\end{abstract}
]

\printAffiliationsAndNotice{\textsuperscript{$\dagger$}Work done while at Adobe. }

\section{Introduction}
\label{sec:introduction}

\begin{figure*}[t]
    \centering
    \includegraphics[width=\linewidth]{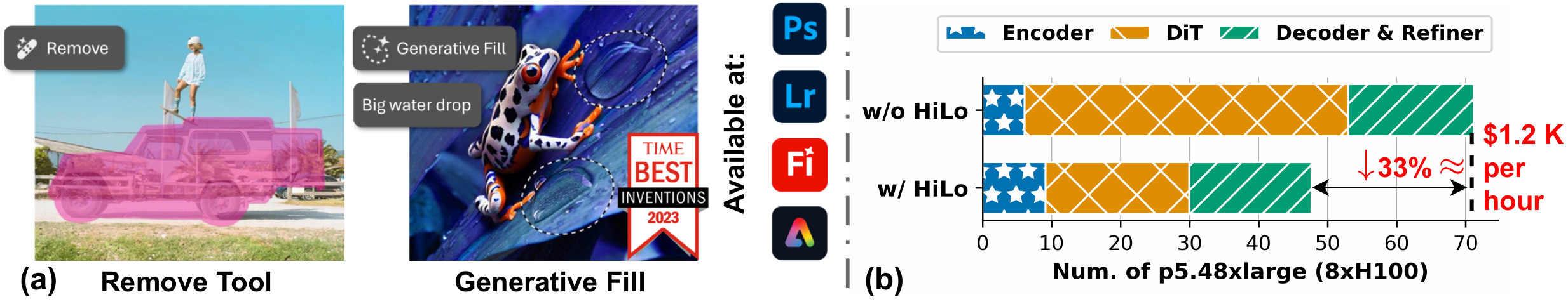}
    \vspace{-1.4em}
    \caption{(a) Representative GenAI features in Adobe products: Remove Tool~\citep{adobe2025remove} and Generative Fill~\citep{adobe2025generativefill}. When no text prompt is provided, Generative Fill invokes the Remove model. (b) Number of Amazon AWS p5.48xlarge nodes used to serve the Remove feature with and without HiLo-Token. The p5.48xlarge node costs \$55.04 per hour based on public data~\citep{vantage2025aws_p548xlarge}.}
    \label{fig:top_figure}
    \vspace{-0.5em}
\end{figure*}

\hr{Creative image editing has received widespread attention in the era of generative artificial intelligence (GenAI). Representative examples include specialist editing models such as the Remove Tool~\citep{adobe2025remove} and Generative Fill~\citep{adobe2025generativefill} in Adobe’s Photoshop, Lightroom, Firefly, and Express, as shown in Fig.~\ref{fig:top_figure} (a), which are fine-tuned from pre-trained generalist Firefly Image models~\citep{adobe2024fireflyimage3,adobe2025fireflyimage4} on carefully curated high-quality data. The space has moved quickly: large-scale generation and editing models from other companies now include Google’s Gemini Image, also known as Nano Banana~\citep{google2025nanobanana,google2025nanobananapro,google2026nanobanana21}, OpenAI’s ChatGPT Image~\citep{openai2025chatgptimages,openai2026chatgptimages25}, ByteDance’s Seedream and SeedEdit~\citep{seedream2025seedream4,wang2025seededit3,bytedance2026seedream5}, Black Forest Labs’ FLUX~\citep{blackforestlabs2025flux2,blackforestlabs2026flux3}, and Alibaba’s Qwen Image~\citep{wu2025qwenimage,yin2025qwenimagelayered}.}

\hr{Across these systems, the dominant emphasis has been on intrinsic model controllability, i.e., inferring which regions to edit or leave untouched from the prompt alone, although recent releases have begun to expose explicit region control as well~\citep{google2026nanobanana21}. Adobe’s specialist and partner models instead operate directly on user-defined masks supplied through Photoshop’s UI, aligning with users’ need for pixel-precise control within established creative workflows. These mask-based workflows are heavily used: within 28 days of the release of Photoshop v27.0 on October 28, 2025, 1.1 million out of 3.3 million users engaged with the Generative Fill feature, resulting in 36.2 million total interactions and 82.8 million generative credits consumed.}

However, serving such GenAI models to millions of users poses significant challenges, particularly as the field transitions from convolution-based U-Nets~\citep{stablediffusion,podell2023sdxl} to Diffusion Transformers (DiTs)~\citep{peebles2023scalable,chen2025unireal,zhang2025uniser} or Transformer-based multimodal models that combine Vision Language Models (VLMs) with DiTs~\citep{adobe2025fireflyimage4,seedream2025seedream4} to achieve higher editing or generation quality. These models are nearly $6\times$ more expensive to serve in cloud environments, driven by both slower DiT inference and the need for hardware upgrades, such as switching from A100 to H100 GPUs, to achieve acceptable latency---even though DiTs have $1.8\times$ fewer parameters than U-Nets.
Existing compression techniques, such as token-level~\citep{bolya2023token,wang2024attention,smith2024todo,chen2024deep} or model-level~\citep{kim2023bk,fang2023structural} compression, activation caching~\citep{xu2018deepcache,moura2019cache,ma2024learning}, and low-bit model quantization~\citep{chen2025q,hwang2025tq}, either provide limited benefits beyond aggressive timestep distillation~\citep{dmd,dmdv2}, which is often the most effective and widely adopted approach in industry, or inevitably introduce quality regression; even small-scale distortions on a small subset of test images can impact many users and are therefore unacceptable for production deployment.
Moreover, only a few prior works, such as LazyDiffusion~\citep{nitzan2024lazy} and DiffCR~\citep{you2025layer}, consider the unique optimization properties of user-defined, mask-based GenAI models in Photoshop UI interaction settings; these methods generate content only within masked regions rather than over the entire image.

\begin{figure*}[t]
    \centering
    \includegraphics[width=0.87\linewidth]{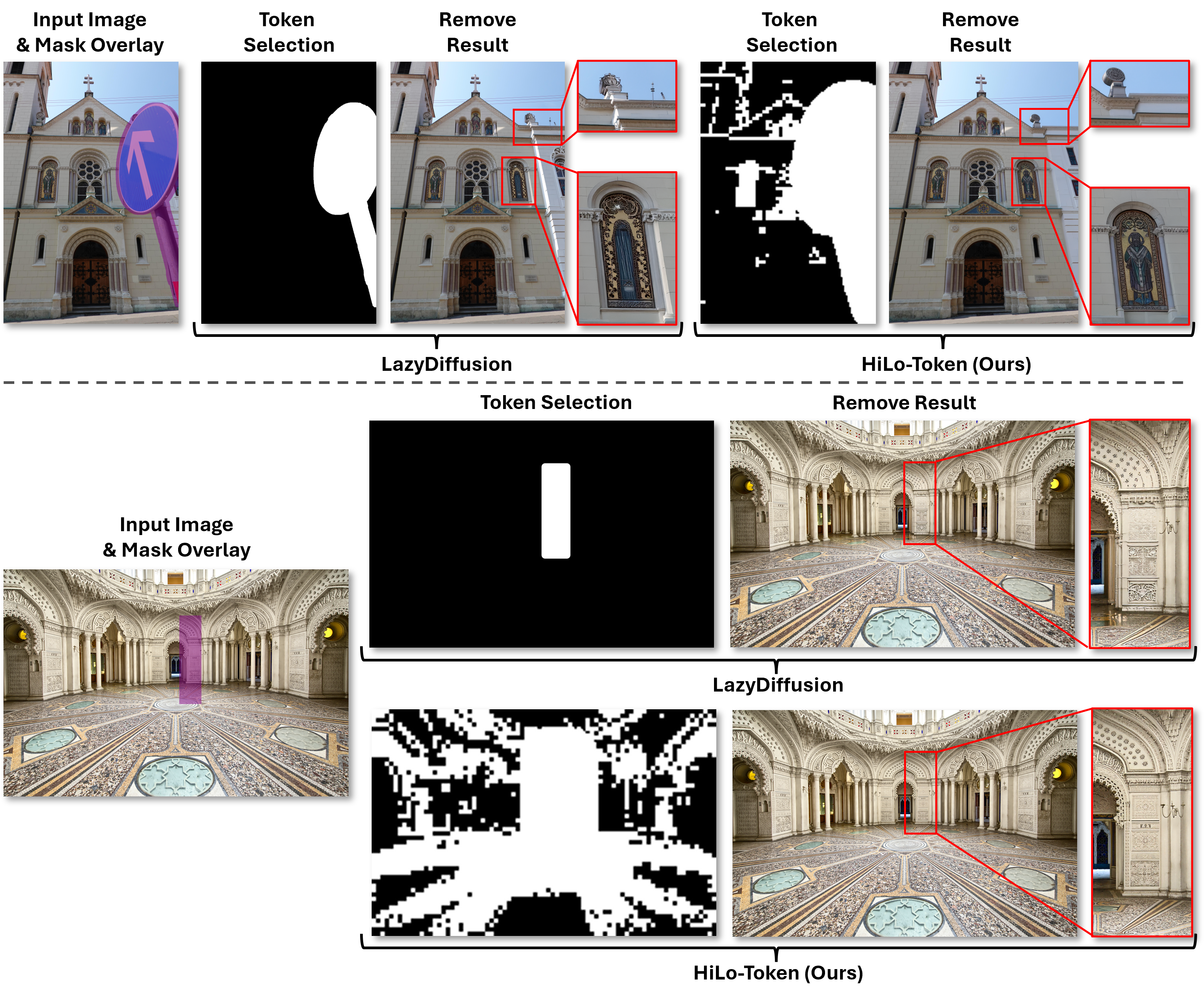}
    \vspace{-1.5em}
    \caption{Two representative examples of images with complex textures or occluded symmetric high-frequency patterns, highlighting the importance of token selection by comparing LazyDiffusion~\citep{nitzan2024lazy} with the proposed HiLo-Token. The purple overlay on each input image marks the user-specified mask, i.e., the region to be removed. In the token selection maps, white regions denote selected tokens and black regions represent pruned tokens. Images are compressed for visualization, with important details zoomed in; the originals are at 30-megapixel (MP) resolution.}
    \label{fig:example}
    \vspace{-1em}
\end{figure*}

To reduce serving costs without quality regression, we present HiLo-Token, an input-adaptive token compression framework that allocates a larger token budget to high-frequency, context-rich regions while assigning fewer tokens to low-frequency areas.
Our HiLo-Token is built on LazyDiffusion~\citep{nitzan2024lazy}, so, by default, we retain all tokens within a dilated mask to preserve strong locality and contextual relevance.
In addition, HiLo-Token addresses three key challenges.
\textit{\textbf{First}, how can editing quality be preserved under aggressive token dropping?} Retaining only tokens within the dilated mask inevitably leads to context loss; even with a separate context encoder, most useful context tokens are discarded to match the mask’s spatial shape, as context and mask tokens are concatenated along the feature dimension~\citep{nitzan2024lazy}. We address this issue through frequency-aware token selection: high-frequency tokens are retained to capture important local details, while tokens from a $16\times$ downsampled image represent low-frequency context that preserves global structure. These additional context tokens are concatenated along the token dimension.
\textit{\textbf{Second}, how can the overhead of obtaining context tokens be reduced?} LazyDiffusion relies on a transformer-based context encoder that must be fully computed even though most tokens are later dropped. We replace this costly model with a lightweight frequency-based selection mechanism that uses only two convolution operations to compute a spatial edge/frequency map for high-frequency token selection, and a single linear layer to patchify downsampled images for low-frequency tokens, adding only a few milliseconds of overhead.
\textit{\textbf{Third}, how can token allocation be input-adaptive?} Image complexity varies widely: simple scenes require minimal context, whereas complex scenes with rich textures or strong structural symmetry relative to the masked regions demand more tokens. Learning-based importance prediction methods, such as those based on attention signals, are costly and unreliable in this setting because (1) relevant content may not yet be generated at early diffusion steps, requiring prediction in the middle of the diffusion process, and (2) they rely on spatial cross-correlation with the masked region, which breaks down when the informative content is occluded or obscured by distractors. For example, when removing a traffic sign in front of a symmetric church painting, as shown in Fig.~\ref{fig:example}, attention-based approaches may overlook the symmetric region on the opposite side, because the matching content inside the masked region is hidden behind the sign and therefore yields no meaningful correlation signal. Instead, we employ robust Sobel-based edge detection to obtain a spatial frequency map, followed by pooling and regionalization, to guide token selection without relying on semantic priors or cross-region relationships.

Extensive experiments on production-level evaluation data validate the effectiveness of the proposed method, which achieves DiT speedups of 3.13$\times$, 2.59$\times$, and 1.67$\times$ on A100-80GB for image editing tasks with small, medium, and large mask ratios, corresponding to average ratios of 6.38\%, 15.92\%, and 35.36\%, respectively, without any regression in generation quality. As shown in Fig.~\ref{fig:top_figure} (b), HiLo-Token reduces the number of Amazon AWS p5.48xlarge nodes required to serve the Remove feature by 33\%.

\section{The Proposed Method}
\label{sec:methods}

\begin{figure*}[t]
    \centering
    \includegraphics[width=0.85\linewidth]{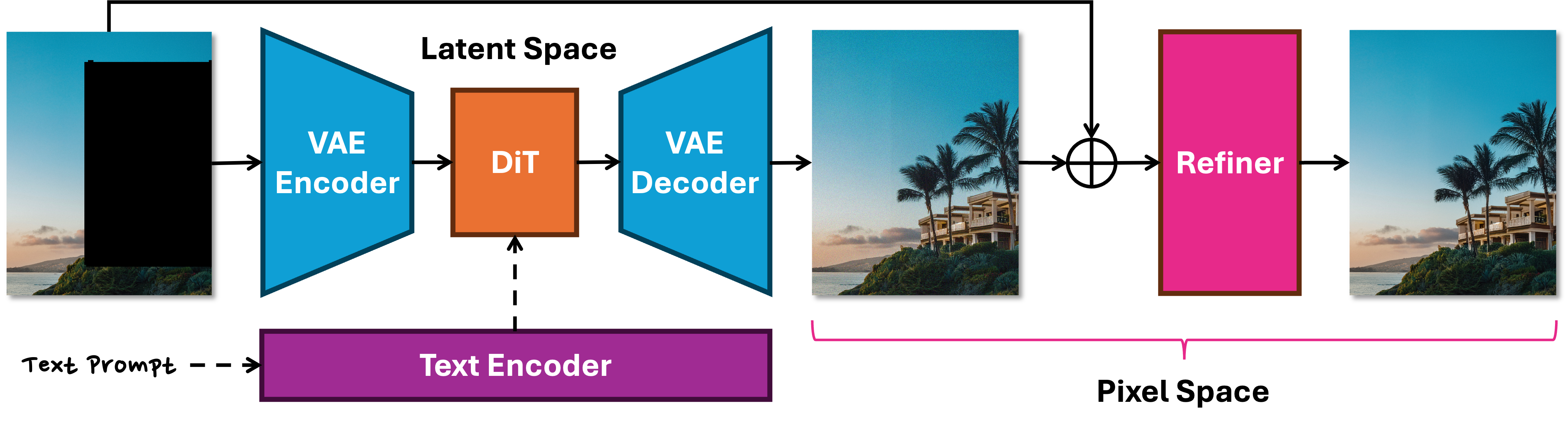}
    \vspace{-1em}
    \caption{Image editing model architecture overview, including a VAE, a DiT, a refiner, and an optional text encoder.}
    \label{fig:me-model}
\end{figure*}

\hr{This section presents the proposed HiLo-Token framework. In Sec.~\ref{sec:me}, we begin with a preliminary description of the image editing generalist model, which is fine-tuned from a multimodal foundation model, Firefly Image 3~\citep{adobe2024fireflyimage3}, and profile it in Sec.~\ref{sec:profiling} to identify the DiT as the dominant latency bottleneck. Guided by an analysis of user mask statistics, we then introduce the HiLo-Token design for input-adaptive token allocation in Sec.~\ref{sec:hilo}. In Sec.~\ref{sec:posttrain}, we detail the post-training procedure and curated data that integrate HiLo-Token into the generalist model and specialize it into task-specific models for Remove, Generative Fill, and Generative Expand. Finally, in Sec.~\ref{sec:serving}, we describe how these specialist models are served in production.}

\begin{figure*}[t]
    \centering
    \includegraphics[width=0.85\linewidth]{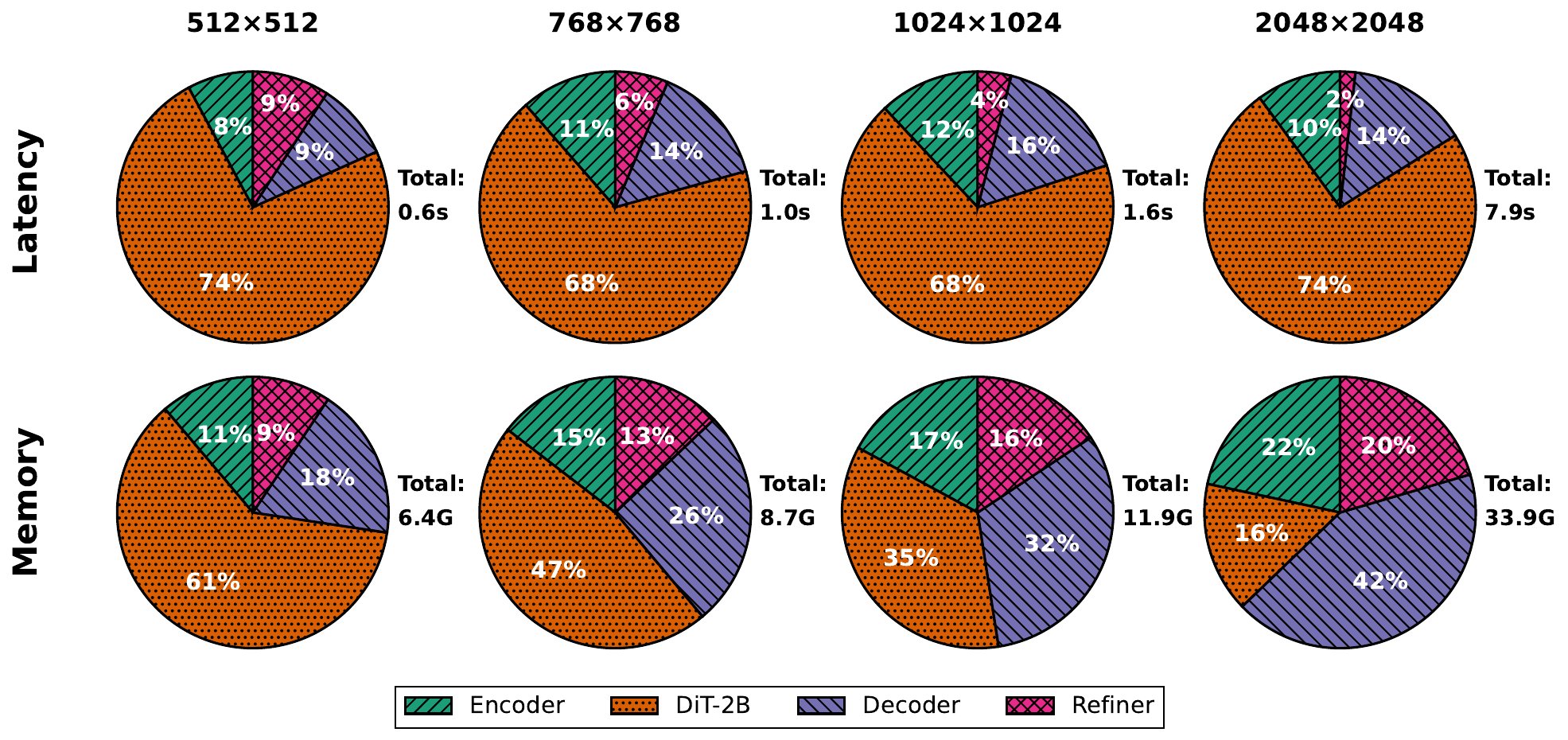}
    \vspace{-1em}
    \caption{Latency and memory profiling results for the image editing model, measured with PyTorch on an A100-80GB GPU across four resolution settings. The VAE encoder latency is counted twice to account for encoding both the image and the mask, while the DiT latency is counted eight times to align with 8-timestep distillation.}
    \label{fig:profiling}
    \vspace{-0.5em}
\end{figure*}

\subsection{Image Editing Generalist Model}
\label{sec:me}

Our HiLo-Token framework is built on top of an image editing generalist model~\citep{chen2025unireal,zhang2025uniser}. The generalist model is fine-tuned from Firefly’s foundation model~\citep{adobe2024fireflyimage3} on a large corpus of image editing data. Note that this is not the largest Firefly Image model; however, it is well suited for image editing deployment due to its uncompromised editing quality at a moderate scale (a 2B-parameter DiT backbone).
It serves as a generalist across a wide range of image editing tasks, including object and effect insertion, removal, replacement, relighting, text editing, camera pose adjustment, and subject extraction. Despite the model's strong general capabilities, certain tasks benefit from dedicated specialist models fine-tuned on carefully curated, task-specific high-quality data, because their objectives may otherwise conflict with those of other tasks (e.g., removal versus insertion).
As a result, the generalist model is further fine-tuned into task-specific specialists, such as Remove and Generative Fill models, which are directly deployed to end users in production.
HiLo-Token is applied during this supervised fine-tuning (SFT) stage.
As illustrated in Fig.~\ref{fig:me-model}, the image editing model, both the generalist and the specialists derived from it, comprises four main components: a VAE encoder, a VAE decoder, a DiT backbone, and a refiner model. A text encoder is optionally included, depending on whether text conditioning is provided. The VAE encodes input images and masks into a latent space, enabling the DiT to operate on compressed latent representations~\citep{stablediffusion}.
The refiner model~\citep{zheng2022image,zheng2025pixperfect} is the only component that operates entirely in pixel space, and it sits closest to the final output; as a result, it is particularly sensitive to compression artifacts. In practice, we use full precision or BF16 for it.

\subsection{Model Profiling}
\label{sec:profiling}

We profile the image editing model to better understand its computational bottlenecks. As shown in Fig.~\ref{fig:profiling}, we report latency and memory profiling results at four commonly used image resolutions: 512$^2$, 768$^2$, 1024$^2$, and 2048$^2$. Among the four components of the image editing model, the DiT module consistently dominates end-to-end latency, accounting for 68--74\% of the total runtime depending on resolution, followed by the VAE decoder, the VAE encoder, and the refiner.
The memory profile exhibits a different trend. Owing to memory-efficient attention mechanisms~\citep{xFormers2022,dao2023flashattention,pytorch_scaled_dot_product_attention}, the DiT module is not the primary memory bottleneck at high resolutions. Its share of total memory decreases from 61\% at 512$^2$ resolution to only 16\% at 2048$^2$. In contrast, the VAE and refiner increasingly dominate memory consumption, becoming the main memory bottlenecks for large-resolution image editing.
\begin{figure*}[t]
    \centering
    \includegraphics[width=0.9\linewidth]{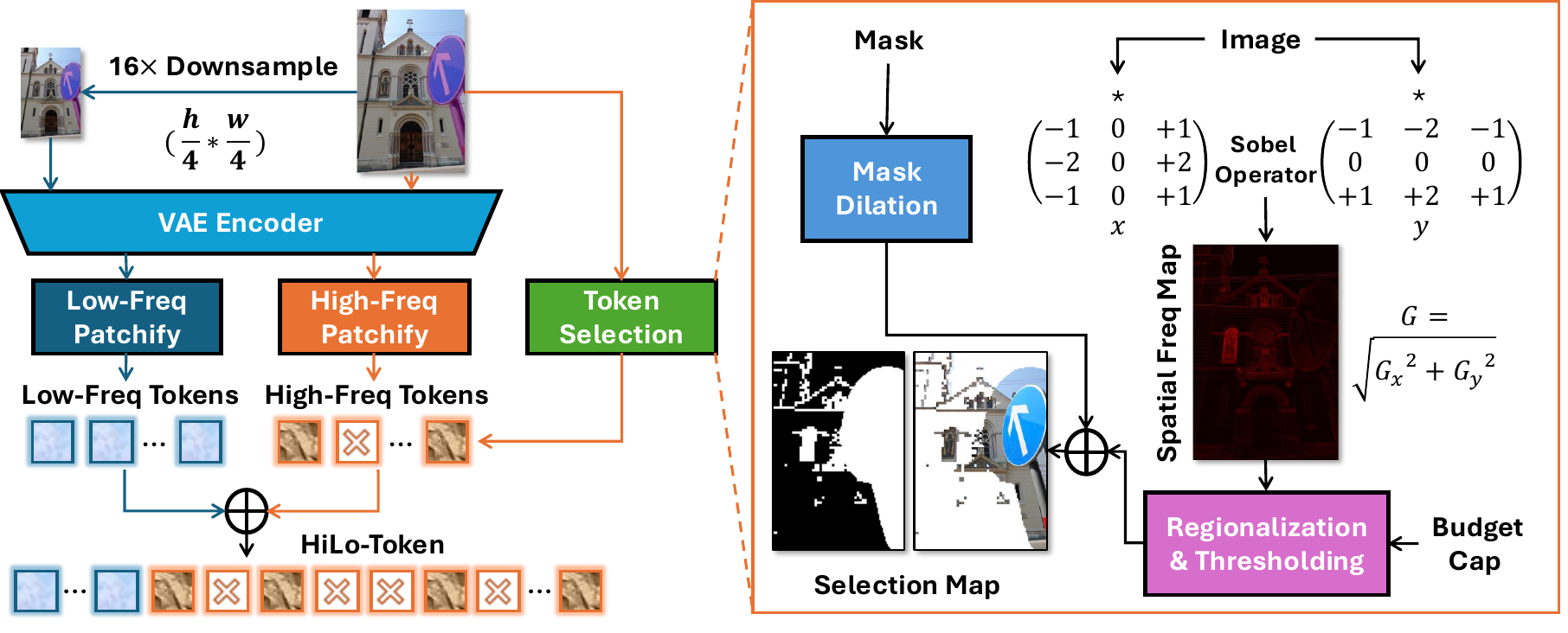}
    \vspace{-1em}
    \caption{Overview of the proposed HiLo-Token framework, illustrating low- and high-frequency token selection.}
    \label{fig:hilo_framework}
    \vspace{-0.5em}
\end{figure*}
For cloud serving on A100 or H100 nodes, latency and throughput are the dominant factors determining cost–benefit efficiency. In contrast, for edge deployment, memory serves as a hard constraint, as consumer GPUs often have limited capacity (e.g., 12 GB), requiring the model to be explicitly optimized to fit within strict memory budgets. In this work, we primarily target latency reduction to lower cloud serving costs; the associated memory reduction of the DiT model is a by-product of this optimization.

\subsection{HiLo-Token Framework}
\label{sec:hilo}

\begin{figure}[t]
    \centering
    \includegraphics[width=\linewidth]{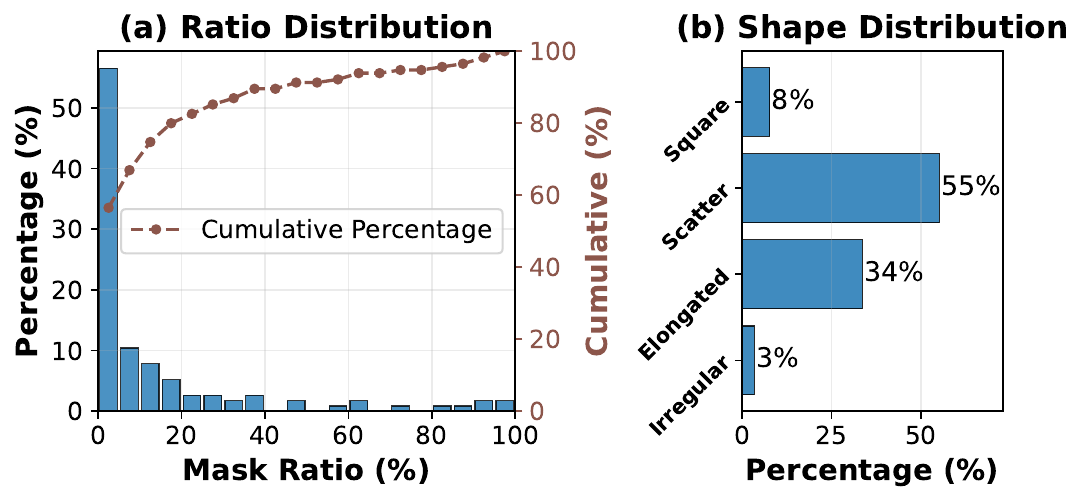}
    \vspace{-1.8em}
    \caption{User editing mask statistics. (a) Mask ratio distribution; (b) Mask shape category distribution.}
    \label{fig:mask_statistic}
    \vspace{-1em}
\end{figure}

\textbf{Mask Distribution.}
Across input images, the user-specified edited regions exhibit substantial diversity. We analyze the mask statistics, as shown in Fig.~\ref{fig:mask_statistic}. In terms of mask ratio, more than 50\% of editing requests involve small masks covering less than 10\% of the image area. We further compute the cumulative distribution (shown as the brown curve), which indicates that in 90\% of cases, users edit no more than 50\% of the image.
Regarding mask shape, the majority of cases (55\%) consist of scattered holes, followed by elongated holes, while the remaining masks are square or irregular.
These statistics suggest that it is unnecessary for the DiT model to operate on the full image in most scenarios. Instead, the content within the masked regions, together with the relevant surrounding context, is sufficient. For example, human retouching tasks typically require processing only the relevant skin regions.

\textbf{HiLo-Token Selection.}
As analyzed above, not all tokens or spatial regions contribute meaningfully to editing results; therefore, careful token selection is crucial for improving efficiency while preserving editing quality. In this work, we propose a HiLo-Token selection method. As illustrated in Fig.~\ref{fig:hilo_framework}, given an input image and the user-provided mask, we employ two parallel branches to extract low-frequency and high-frequency tokens.
For low-frequency tokens, we aggressively downsample the input image by 16$\times$ and then pass it through the VAE encoder followed by a low-frequency patch-embedding linear layer to capture coarse global structure. Because this aggressive downsampling yields a very small number of tokens, we retain all low-frequency tokens.
For high-frequency tokens, we avoid aggressive downsampling and use a VAE with a moderate 8$\times$ compression ratio, together with a patch-embedding layer using an additional 2$\times$ compression, to preserve spatial resolution and prevent over-smoothing in the generated output. As a result, this branch produces a large number of tokens, making effective token selection essential.
While attention-based methods can provide guidance for token selection, they rely on correlation priors that may fail under occlusions. For example, when removing a traffic sign occluding a symmetric church painting, attention-based approaches may overlook the symmetric region on the opposite side, because the matching content inside the masked region is hidden behind the sign and therefore yields no meaningful correlation signal. To avoid such limitations, we adopt a simple, correlation-free strategy: we directly estimate a normalized spatial-frequency map using Sobel operators and select tokens whose frequency magnitude exceeds a threshold (e.g., 0.1).
However, naive thresholding yields spatially scattered tokens that the model cannot process efficiently. To address this, we further regionalize the selected token map by applying 16$\times$ spatial pooling to the frequency map, aligning it with the token grid. This produces more coherent, region-level token selections. In addition, we dilate the user mask to retain nearby contextual tokens and combine the dilated mask with the high-frequency selection map.
The computational cost of this token selection is negligible (approximately 10 ms).
Finally, the selected high-frequency tokens are concatenated with the low-frequency tokens to construct the HiLo-Token representation, which is subsequently processed by the DiT.
We note that the proposed input-adaptive HiLo-Token representation generalizes well to a broad spectrum of editing tasks and the associated generalist or specialist models.

\subsection{Post-training with HiLo-Token}
\label{sec:posttrain}

\textbf{Supervised Fine-Tuning (SFT).}
The first step in adapting the generalist into a specialist for image editing tasks such as Remove or Generative Fill is to perform supervised fine-tuning on a carefully curated dataset, with HiLo-Token token selection active throughout.
For instance, when fine-tuning the Remove model, we use 407,630 image–mask pairs at multiple image resolutions, spanning datasets for object removal (including both synthetically rendered and real-world data), retouching, object stitching and composition, manual masking, and omni-edit mixed tasks.
Fine-tuning is initialized from the generalist checkpoint and runs for approximately 20K iterations. We observe that the fine-tuning recipe must be carefully co-designed with the data distribution. In particular, if too few object-removal samples are included, the model requires significantly longer training to suppress object insertion artifacts during removal. Conversely, excessively long fine-tuning can exacerbate seam artifacts, so the schedule must be monitored carefully to balance the two.
Moreover, different editing tasks inherently favor different fine-tuning recipes to achieve optimal quality. Generative Fill and Remove, for example, are trained as separate specialist models, since these tasks are partially contradictory: one focuses on adding content while the other removes it. Joint optimization can lead to interference effects, such as insertion artifacts during removal, motivating task-specific fine-tuning strategies.

\textbf{Few-step Distillation.}
After obtaining the teacher model via SFT, we perform distillation to improve inference efficiency prior to deployment. The teacher model employs a 50-step diffusion process, which is prohibitively slow for practical use, requiring approximately 7~seconds to generate a 1K-resolution image even on high-end A100 GPUs.
Ideally, we would like the fast student generator to produce samples that are indistinguishable from those generated by the teacher. Following Distribution Matching Distillation (DMD)~\citep{dmd,dmdv2}, we minimize the Kullback--Leibler (KL) divergence between the student and teacher image distributions, denoted as $p_{\text{s}}$ and $p_{\text{t}}$, respectively:
\begin{align*}
D_{\mathrm{KL}}\!\left(p_{\text{s}} \,\|\, p_{\text{t}}\right)
&= \mathbb{E}_{x \sim p_{\text{s}}}
\left(
\log \frac{p_{\text{s}}(x)}{p_{\text{t}}(x)}
\right) \\
&= \mathbb{E}_{z \sim \mathcal{N}(0, I)}
\left(
\log p_{\text{s}}(x) - \log p_{\text{t}}(x)
\right),
\end{align*}
where $x = G_{s}(z)$.
Directly computing the likelihood term is intractable. Following DMD, we optimize this objective via gradient-based distribution matching using the score functions of the teacher and student models. In practice, the teacher provides an estimate of the score $\nabla_x \log p_{\text{t}}(x)$, which serves as a supervision signal for aligning the student distribution with that of the teacher. This formulation effectively transfers generative knowledge from the teacher to the student, enabling few-step (e.g., 8-step) inference while preserving high-fidelity editing quality.

\begin{figure}[t]
    \centering
    \includegraphics[width=\linewidth]{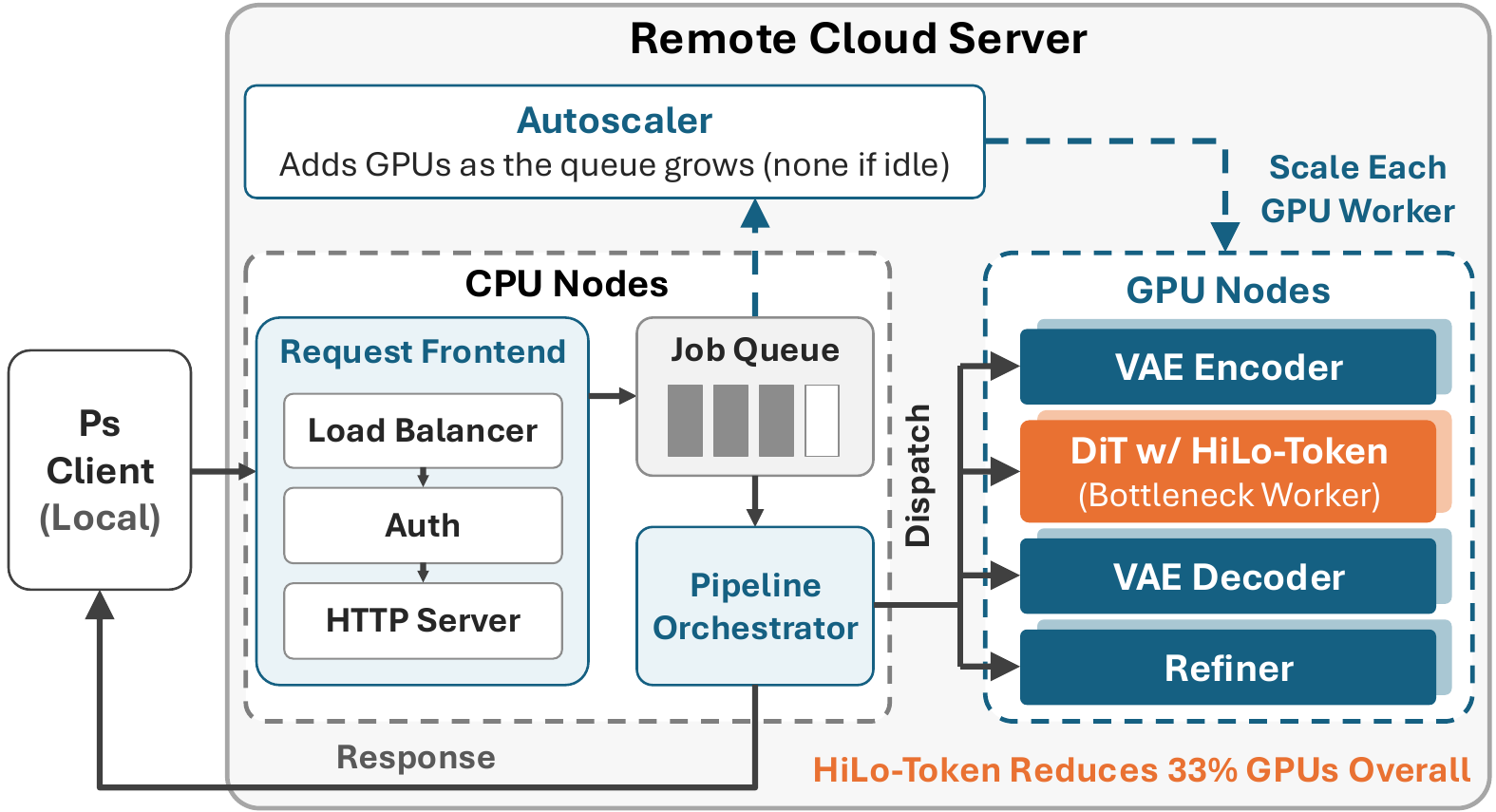}
    \vspace{-1.5em}
    \caption{\hr{Serving system architecture of the deployed editing models. Local Photoshop clients send edit requests to a remote cloud server, where CPU nodes run the request frontend and place jobs on a queue; a pipeline orchestrator then dispatches sub-tasks to the GPU nodes and returns responses to the clients. Each GPU worker, i.e., the VAE encoder, the DiT, the VAE decoder, and the refiner, is replicated and scaled independently, with the autoscaler adding GPU replicas as the queue grows and releasing them when idle. Because the DiT is the bottleneck worker, accelerating it with HiLo-Token reduces the overall GPU count by 33\%.}}
    \label{fig:serving_arch}
    \vspace{-1.5em}
\end{figure}

\subsection{HiLo-Token Serving System}
\label{sec:serving}

\hr{\textbf{Overview.}
Fig.~\ref{fig:serving_arch} summarizes how the deployed models are served. Edit requests from the Photoshop client are admitted by a CPU frontend, queued, and dispatched by a pipeline orchestrator to a set of disaggregated GPU workers that each execute one stage of the model pipeline. The design is organized around a single constraint: GPU capacity dominates serving cost, so the system is built to keep expensive GPU workers busy under load and to release them as soon as they are idle. HiLo-Token's token reduction applies exactly to the bottleneck DiT GPU workers, as detailed below.}

\hr{\textbf{Disaggregated Model Serving.}
Each stage of the editing pipeline, i.e., the VAE encoder, the DiT backbone, the VAE decoder, and the refiner, is deployed as an independently scheduled worker rather than bundled into a single monolithic process. This prevents a slow stage from stalling the rest of the pipeline, accommodates the very different workload profiles of the individual components, and allows requests to be batched continuously within each stage.
To keep queuing cheap, only a small job descriptor, rather than the image payload itself, traverses the queue; the orchestrator fetches the payload and dispatches sub-tasks to the GPU workers, passing intermediate tensors between stages. This disaggregation is deliberate: GPU capacity is by far the dominant cost, so isolating the GPU-bound DiT stage into its own worker type lets it be scaled, batched, and scheduled independently from the lighter VAE and refiner stages and from the CPU-only orchestration tier, which runs on inexpensive commodity instances.}

\hr{\textbf{Autoscaling.}
Because the DiT worker is both the most expensive and the most latency-critical component (Sec.~\ref{sec:profiling}), it is autoscaled from a real-time queue-depth signal rather than a fixed replica count, including scaling down to zero GPU replicas during idle periods to eliminate cost, at the expense of a cold-start penalty (reloading model weights) on the first request after idling. A predictive scaling layer forecasts near-term traffic to pre-warm GPU replicas ahead of anticipated demand, trading a small amount of idle GPU time for reduced cold-start-induced tail latency.}

\hr{\textbf{Production Deployment.}
The serving stack is replicated across multiple cloud regions for redundancy, and changes are promoted through a progression of environments (development, staging, pre-release, production) for safe incremental rollout. Production traffic for all tasks is further spread across GPU hardware generations (NVIDIA A10, A100, and H100), with each worker's GPU memory requirement automatically constraining it to compatible instance types.
Because HiLo-Token accelerates only the DiT stage, its benefit is amplified rather than diluted by this disaggregated design: the DiT worker is simultaneously the end-to-end latency bottleneck (Sec.~\ref{sec:profiling}, 68--74\% of runtime) and the most expensive unit to autoscale. A faster DiT forward pass directly shortens GPU occupancy per request, letting each GPU replica serve more requests before the autoscaler must provision another GPU worker, which is the underlying mechanism behind the 33\% reduction in AWS GPU nodes required to serve the Remove feature reported in Fig.~\ref{fig:top_figure} (b).}

\section{Experiments}
\label{sec:experiments}

\subsection{Experiment Settings}

\textbf{Model and Dataset.}
We apply our HiLo-Token framework on top of the image editing generalist model described in Sec.~\ref{sec:me}. For SFT, we use an internally curated dataset comprising 407,630 image–mask pairs, spanning multiple editing categories and including both synthetic and real data.

\textbf{SFT, Distillation, and Inference Settings.}
For SFT, we optimize the DiT and patchify layers using AdamW with a learning rate of $1.2\times 10^{-5}$, weight decay $10^{-2}$, and $\beta=(0.9,0.95)$, together with a linear warmup–cosine decay schedule with 2K warmup steps and a minimum learning rate of $10^{-5}$. Training is performed with fully sharded data parallel (FSDP) across four nodes of $8\times$A100 GPUs, gradient clipping set to $1.0$, an exponential moving average (EMA) with decay $0.9999$, and checkpoints saved every 1K steps.
For DMD-style distillation, we train an 8-step student using the timestep list $\{999,917,825,724,612,486,344,184\}$ under BF16 mixed precision and FSDP across four nodes of $8\times$A100 GPUs, using AdamW with a learning rate of $2\times 10^{-5}$ and weight decay $10^{-2}$, a total batch size of 128, EMA decay $0.995$ from step 0, and a DMD objective augmented with adversarial learning with a GAN weight of 0.1. The discriminator is trained with AdamW using a learning rate of $5\times 10^{-5}$ and a softplus loss.
For inference, we evaluate on a suite of internal benchmarks, generating four samples per input; sampling uses the same timestep list as distillation, a constant classifier-free guidance (CFG) scale of 1.0, a prompt suffix \texttt{\#photorealistic}.
% and a negative prompt \texttt{cartoonish}, \texttt{distorted}, \texttt{mask}, \texttt{deformed}, \texttt{low quality}, \texttt{poor aesthetics}.
%
We exclude samples containing harmful or biased content and restrict evaluation to prevent malicious use cases, such as attempts to remove clothing or otherwise violate personal integrity.

\textbf{Evaluation.} \hr{Existing quantitative metrics, such as FID and CLIP score, are unreliable proxies for production editing quality: it is straightforward to construct edits with a visible defect, such as a seam, a duplicated object, or a badly blended region, that nonetheless score favorably, because these metrics summarize global distributional or embedding-level similarity rather than judging local, task-specific correctness. Shipping a consumer feature requires catching exactly these long-tail failures, which aggregate statistics silently average away. We therefore treat evaluation by a dedicated quality engineering (QE) team, who compare results with and without HiLo-Token across a wide range of categories, as the source of truth for all quality comparisons.} Latency improvements are measured on A100-80GB GPUs.

\subsection{Generalist Performance}

The generalist model supports a wide range of image editing tasks with consistent performance. On the ImgEdit benchmark~\citep{ye2025imgedit}, it achieves an overall score (4.3) comparable to state-of-the-art models such as FLUX Kontext~\citep{labs2025flux}, Nano Banana~\citep{google2025nanobanana}, and GPT Image~\citep{openai2025chatgptimages}.
Moreover, on soft effect removal tasks, the generalist model surpasses all competing models and methods~\citep{zhang2025uniser}.

\subsection{Result of Specialists with HiLo-Token}

\begin{figure*}[t]
    \centering
    \includegraphics[width=\linewidth,height=0.85\textheight,keepaspectratio]{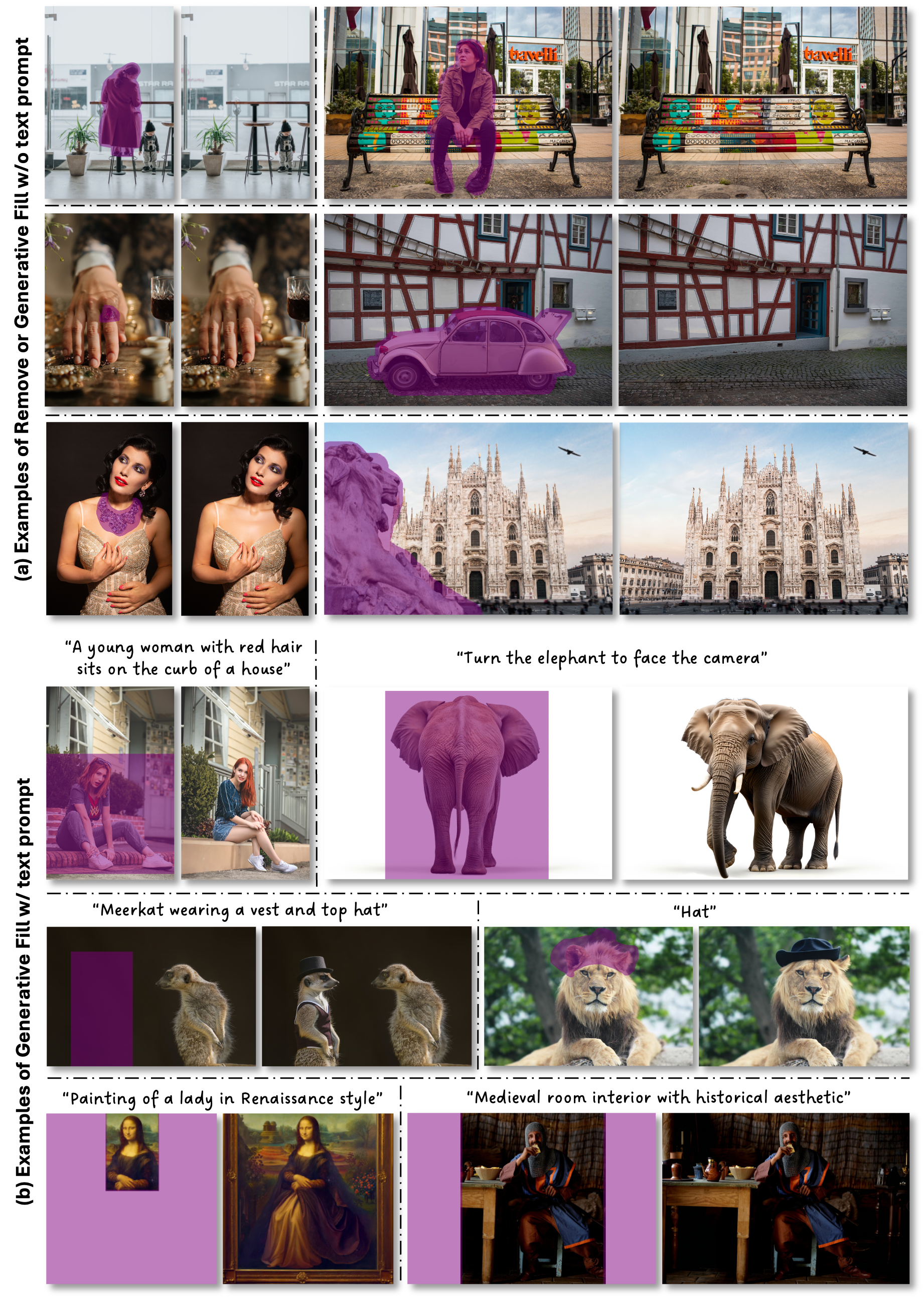}
    \vspace{-0.5em}
    \caption{Representative examples of the Remove Tool and Generative Fill with our token compression applied.}
    \label{fig:more_example}
    \vspace{-1em}
\end{figure*}

\begin{figure}[t]
    \centering
    \includegraphics[width=\linewidth]{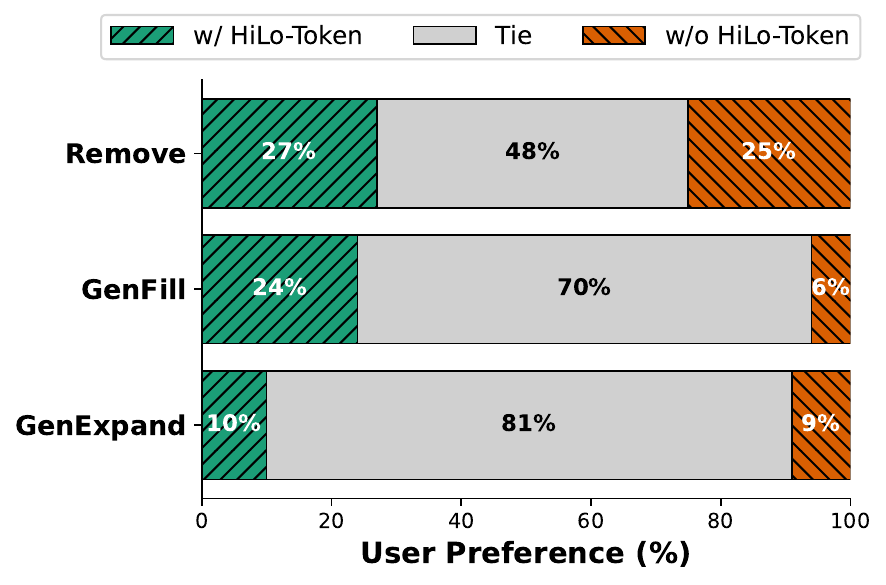}
    \vspace{-2em}
    \caption{User study results w/ and w/o HiLo-Token.}
    \label{fig:user_study}
    \vspace{-1.5em}
\end{figure}

\textbf{Qualitative Results and User Study.}
\hr{Fig.~\ref{fig:more_example} shows representative outputs of the Remove Tool and Generative Fill with HiLo-Token applied.} To quantify these observations, we conduct a user study comparing model performance with and without the proposed HiLo-Token. \hr{Three quality engineers independently score blind, randomized A/B comparisons over several hundred edited images per task as a win, tie, or loss.} As shown in Fig.~\ref{fig:user_study}, we evaluate three specialist models for Remove, Generative Fill, and Generative Expand, all of which are available in Adobe products. In most cases, token pruning produces editing results of comparable quality, with tie rates of 48\%, 70\%, and 81\% for the three tasks, respectively. Models w/ and w/o HiLo-Token perform similar, with comparable numbers of winning cases, and in some settings the model with HiLo-Token performs better. Notably, for Generative Fill with text prompts, tokens outside a dilated mask can be more aggressively or even fully pruned, since the inserted content is largely self-contained within the user-specified mask area.

\hr{\textbf{Comparison with LazyDiffusion.}
\begin{table}[t]
\centering
\caption{\hr{Editing quality relative to the uncompressed model on 111 representative cases, scored with the same blind A/B protocol.}}
\label{tab:lazy_comparison}
\begin{small}
\resizebox{\linewidth}{!}{
\begin{tabular}{l|c|c|c}
\toprule
Variant & Uncompressed wins & Variant wins & Tie \\
\midrule
LazyDiffusion & 20.7\% & 5.4\% & 73.9\% \\
HiLo-Token & 2.7\% & 3.6\% & 93.7\% \\
\bottomrule
\end{tabular}
}
\end{small}
\vspace{-1em}
\end{table}

Discarding every token outside the dilated mask, as in LazyDiffusion~\citep{nitzan2024lazy}, is faster than HiLo-Token but not shippable. Table~\ref{tab:lazy_comparison} scores both variants against the uncompressed model under the same A/B protocol over 111 comparisons: LazyDiffusion loses 20.7\% of cases and wins only 5.4\%, a clearly visible regression. Restoring frequency-aware context with HiLo-Token lifts the tie rate from 73.9\% to 93.7\% and balances the remainder (3.6\% wins against 2.7\% losses), reaching parity with the uncompressed model. HiLo-Token therefore trades a little of LazyDiffusion's speed for the quality headroom that deployment requires.}

\begin{figure}[t]
    \centering
    \includegraphics[width=\linewidth]{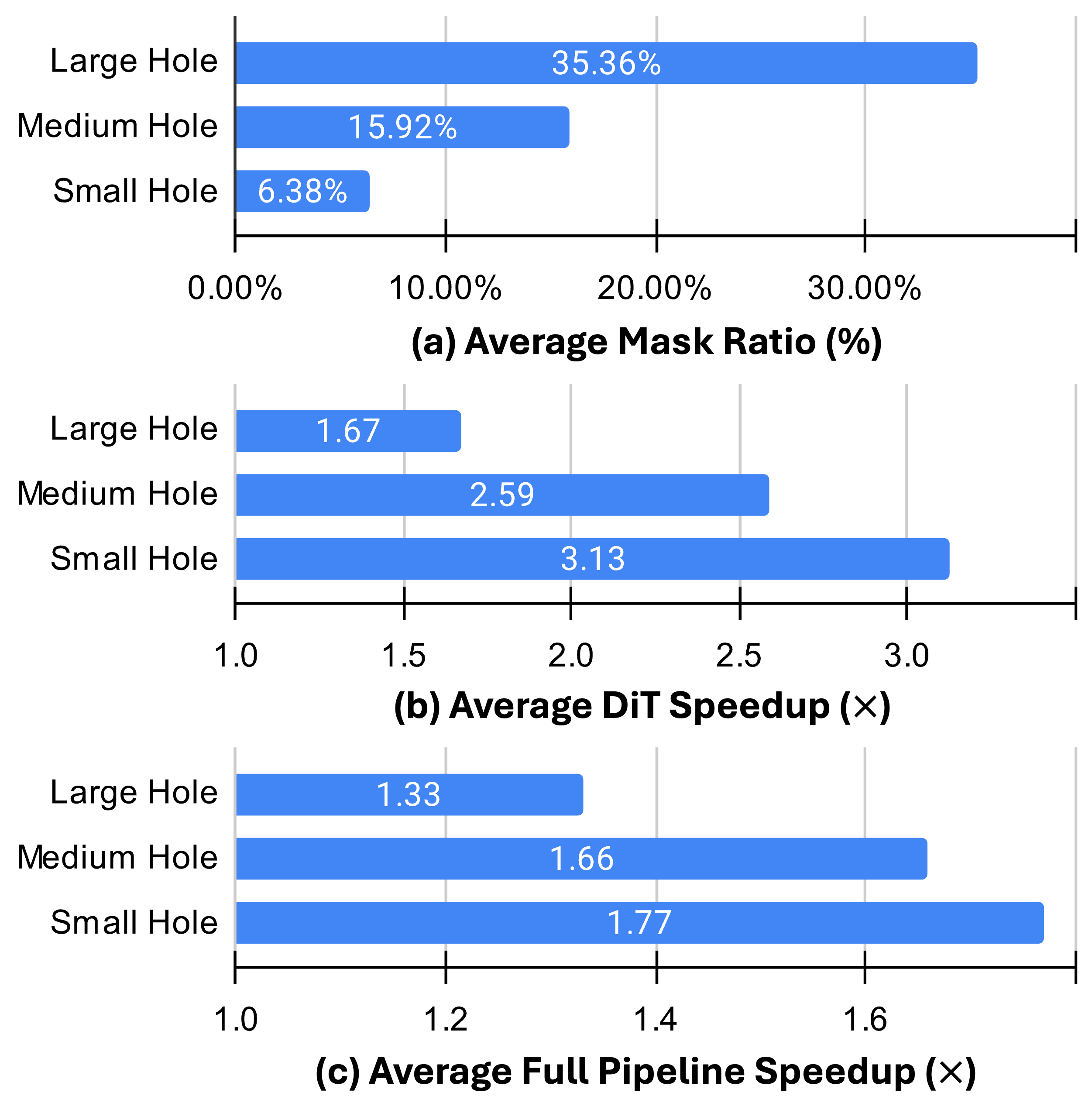}
    \vspace{-2em}
    \caption{Latency speedup results for the editing model with HiLo-Token across three mask-ratio groups.}
    \vspace{-1.5em}
    \label{fig:speedup}
\end{figure}

\textbf{Speedup Analysis.}
We evaluate the latency speedup achieved by applying HiLo-Token. Specifically, we measure performance on 92 representative user editing cases and partition them into three evenly sized groups based on mask ratio, corresponding to small, medium, and large holes, with average mask ratios of 6.38\%, 15.92\%, and 35.36\%. As shown in Fig.~\ref{fig:speedup}, HiLo-Token accelerates the DiT by $3.13\times$, $2.59\times$, and $1.67\times$ for the three groups, respectively: smaller masks retain fewer tokens and therefore benefit most. Over the full inference pipeline, these translate to overall speedups of $1.77\times$, $1.66\times$, and $1.33\times$. These results demonstrate the effectiveness of our HiLo-Token.

\begin{figure*}[t]
    \centering
    \includegraphics[width=0.315\linewidth]{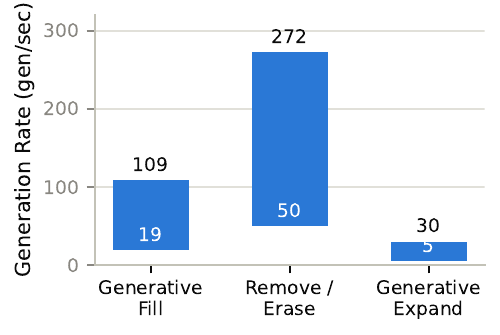}%
    \hspace{0.015\linewidth}%
    \includegraphics[width=0.325\linewidth]{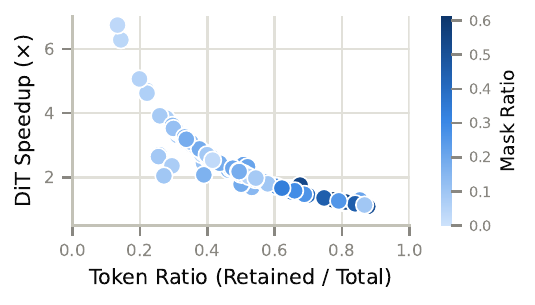}%
    \hspace{0.015\linewidth}%
    \includegraphics[width=0.325\linewidth]{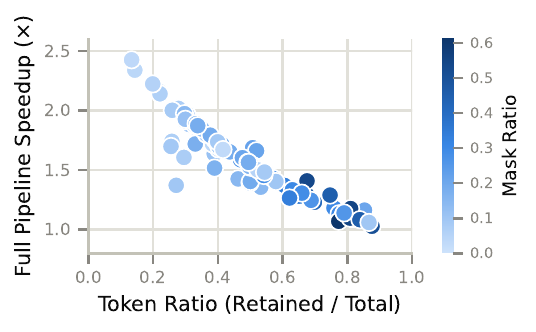}
    \vspace{-1em}
    \caption{\hr{(a) Generation rate range (low to peak traffic) per editing feature in Photoshop, from a representative telemetry window; Remove accounts for the largest share, served either via the Generative Fill button with an empty prompt or the Remove button directly. (b, c) DiT-only and full-pipeline speedup versus retained token ratio over the 92 production cases, colored by mask ratio. Only the DiT is token-compressed, so its gains are diluted across the full six-stage pipeline. Speedup grows roughly inversely with token ratio, and mask ratio predicts the resulting token budget (Pearson correlation $r=0.76$), motivating input-adaptive rather than fixed-ratio allocation.}}
    \label{fig:throughput}
    \vspace{-1em}
\end{figure*}

\hr{\textbf{Production Serving Benchmark.}
Averaged directly over the same 92 cases, HiLo-Token reduces mean per-case DiT latency from 22.40s to 10.57s ($2.46\times$) and mean full end-to-end pipeline latency from 27.63s to 17.02s ($1.59\times$), consistent with the bucketed averages above. These per-request gains matter at fleet scale: live serving telemetry shows sustained generation rates ranging from 19--109, 50--272, and 5--30 generations/sec for Generative Fill, Remove, and Generative Expand, respectively (Fig.~\ref{fig:throughput} (a)), so the per-case latency reduction translates directly into serving-cost savings under real production load.}

\hr{\textbf{Ablation Study: Downsampling Ratio and Threshold.}
We sweep the low-frequency branch's per-side downsampling factor over $2\times$, $4\times$, and $6\times$ ($4\times$, $16\times$, and $36\times$ fewer pixels). We adopt $4\times$ per side ($16\times$ in area): quality matches the uncompressed baseline while the branch contributes few tokens. The milder $2\times$ also preserves quality but retains four times as many tokens, giving up much of the saving, whereas raters judge $6\times$ worse than $4\times$ on 90\% of A/B comparisons due to over-smoothing. The threshold is far less sensitive: perturbing it around 0.1 shifts the win/tie/loss statistics by at most 2\%, so we fix it at 0.1.}

\hr{\textbf{Ablation Study: Token Ratio Drives Speedup.}
Fig.~\ref{fig:throughput} (b, c) relates token retention to latency at the per-case level. Speedup correlates strongly with token ratio (Pearson correlation $r=-0.86$ for DiT-only) and only moderately with mask ratio alone ($r=-0.62$), confirming that mask ratio is a useful but imperfect proxy: at a fixed mask ratio, the required token budget still varies with scene complexity, which is why HiLo-Token allocates by frequency content rather than mask size. We also tried fixed-ratio Top-K selection in place of the frequency threshold of Sec.~\ref{sec:methods}; on textured or symmetric images it produced scattered, non-contiguous tokens that degraded quality and required per-scene retuning, validating our regionalized, threshold-based design.}

\textbf{Compatibility with Other Optimizations.}
HiLo-Token is fully compatible with quantization techniques, such as FP8, yielding up to 40\% additional latency reduction on DiTs.
It also integrates seamlessly with timestep distillation: while our main results use 8 steps to preserve optimal quality, the model can be further distilled to 5 steps, achieving an additional 37.5\% latency reduction with only minor and acceptable quality degradation affecting fewer than 5\% of images.
Moreover, HiLo-Token remains compatible with optimizations applied to the VAE and refiner models.

\hr{\textbf{Limitations.} HiLo-Token's frequency-based token selection is deliberately conservative. On images with dense fine-grained texture spread across the frame (e.g., foliage, fabric, or heavily textured backgrounds), the Sobel-based frequency map can flag nearly all of the region outside the dilated mask as high-frequency, so the retained token budget and the resulting speedup degrade toward that of the uncompressed baseline on such images. We experimented with more aggressive thresholds to recover speedup in these cases, but found that the resulting quality loss, though confined to a small fraction of edge-case images, was not acceptable: the bar for shipping a regression in a consumer product is considerably higher than the bar for reporting an average improvement in an academic benchmark. We therefore tune HiLo-Token to fail safe, treating reduced speedup on complex images as an acceptable cost of guaranteeing no visible quality regression.}

\section{Related Work}
\label{sec:related_works}

\textbf{Efficient DiTs.} 
DiTs~\citep{peebles2023scalable} are computationally expensive, primarily due to attention's quadratic complexity in the number of tokens. A growing body of work therefore targets their deployment efficiency.
Existing approaches can be broadly categorized along three orthogonal dimensions: token, layer, and timestep optimization.
Token-level methods reduce the number of tokens involved in attention. Prior work explores token merging~\citep{bolya2023token}, token pruning~\citep{wang2024attention,whalen2025early,kong2025token}, and spatial resolution downsampling~\citep{smith2024todo} to eliminate redundant tokens. Task-specific strategies~\citep{nitzan2024lazy,you2025layer} exploit sparsity patterns specific to inpainting.
Layer-level efficiency reduces computation across network depth, via layer pruning~\citep{kim2023bk}, channel pruning~\citep{fang2023structural}, and intermediate feature caching~\citep{xu2018deepcache,moura2019cache,ma2024learning}. However, these methods often incur quality degradation and may be less effective when combined with timestep distillation.
Timestep-level efficiency targets the iterative nature of sampling; knowledge distillation~\citep{salimans2022progressive,meng2023distillation,dmd,kang2024distilling,dmdv2,sauer2025adversarial} sharply reduces the required steps.
Our HiLo-Token instead targets input-adaptive token pruning that improves efficiency without quality regression on image editing tasks.

\textbf{Dynamic Inference.}
Prior work on improving inference efficiency falls into static model compression~\citep{deng2020model} and dynamic inference strategies that adapt computation based on the input or intermediate states~\citep{wang2020dual,zhao2024dynamic,wu2018blockdrop,wang2018skipnet,raposo2024mixture}. Dynamic methods typically modulate execution along network depth, e.g., via early exits~\citep{teerapittayanon2016branchynet,huang2017multi,moon2023early,you2019drawing} or conditional layer skipping, often relying on auxiliary predictors or gating networks. Finer-grained adaptations act at the channel or token level, such as channel skipping~\citep{mullapudi2018hydranets} and mixture-of-depths models~\citep{raposo2024mixture,you2025layer}, which allow different tokens to traverse different computational paths.
Our HiLo-Token focuses on input-adaptive token selection tailored to DiTs, enabling dynamic computation for each image editing query while preserving editing quality. \hr{Distinct from these largely benchmark-driven studies, HiLo-Token is deployed in a large-scale production system, serving millions of user editing requests in Adobe products.}

\section{Conclusion}
\label{sec:conclusion}

In this work, we propose HiLo-Token, an input-adaptive and principled token compression framework that allocates more computation to high-frequency, context-rich regions while assigning fewer tokens to low-frequency areas.
HiLo-Token is used to train efficient generative image editing models that are deployed in Adobe applications, such as Generative Fill and Remove features.
By enabling dynamic token allocation without compromising editing quality, HiLo-Token provides a scalable solution for accelerating DiTs in real-world creative workflows.
Users can experience these models directly in the latest Photoshop to assess the generation quality.

\section*{Acknowledgements}

We acknowledge the scientists, engineers, and leaders across Adobe Firefly and Research who built Firefly Image 3 and the associated generalist editing model, as well as the many engineers who helped ship the feature, such as Shayan Chandrashekar. We also thank the QE teams for their evaluation efforts, including Yukie Takahashi, Pablo Serrano, Rick Mandia, and Irina Maderych. We are grateful to our product managers, Meredith Payne Stotzner and Dongmei Li, for their coordination and support. Finally, we acknowledge the leadership from the DI organization, including Sohrab Amirghodsi, Betty Leong, Sarah Kong, David Hackel, and Maria Yap.

\section*{AI Usage Disclosure}

We used Claude Code (Anthropic) as a writing assistant, limited to grammar and wording edits, condensing author-written text to meet the page limit, and LaTeX formatting.
We also used Claude Code and Cursor to refine plotting scripts that visualize our measured results; all plotted data come from our own experiments.
No generative AI tools were used for ideation, literature search, experiment design, data generation, or code for the proposed method and its experiments.
Edited images in this paper are outputs of the evaluated models, shown as experimental results; all input images are either openly licensed or captured in-house.
All AI-assisted edits were reviewed by the authors, and all technical content, claims, and results were written and verified by the authors, who take full responsibility for all content of this work.
\FloatBarrier

\bibliography{ref}

@misc{adobe2025remove,
      title={Remove Tool in Adobe Photoshop},
      author={{Adobe}},
      year={2025},
      howpublished = {\url{https://www.adobe.com/products/photoshop/remove-object.html}},
      note={Adobe Product}
}

@misc{adobe2025generativefill,
      title={Generative Fill: AI-Powered Image Editing in Adobe Photoshop},
      author={{Adobe}},
      year={2025},
      howpublished = {\url{https://www.adobe.com/products/photoshop/generative-fill.html}},
      note={Adobe Product}
}

@misc{adobe2024fireflyimage3,
      title={Adobe Introduces Firefly Image 3 Foundation Model to Take Creative Exploration and Ideation to New Heights},
      author={{Adobe}},
      year={2024},
      month={April},
      day={23},
      howpublished = {\url{https://news.adobe.com/news/news-details/2024/adobe-introduces-firefly-image-3-foundation-model-to-take-creative-exploration-and-ideation-to-new-heights}},
      note={Adobe Newsroom}
}

@misc{adobe2025fireflyimage4,
      title={Adobe Firefly: The next evolution of creative AI is here},
      author={{Adobe}},
      year={2025},
      month={April},
      day={24},
      howpublished = {\url{https://blog.adobe.com/en/publish/2025/04/24/adobe-firefly-next-evolution-creative-ai-is-here}},
      note={Adobe Blog}
}

@misc{openai2025chatgptimages,
  title        = {The new ChatGPT Images is here},
  author       = {{OpenAI}},
  year         = {2025},
  month        = dec,
  day          = {16},
  howpublished = {\url{https://openai.com/index/new-chatgpt-images-is-here/}},
  note         = {OpenAI Product Announcement}
}

@misc{openai2026chatgptimages25,
  title        = {Introducing ChatGPT Images 2.5},
  author       = {{OpenAI}},
  year         = {2026},
  month        = sep,
  day          = {8},
  howpublished = {\url{https://openai.com/index/introducing-chatgpt-images-2-0/}},
  note         = {OpenAI Product Announcement; API models \texttt{gpt-image-2.5}}
}

@misc{google2026nanobanana21,
  title        = {Nano Banana 2.1: Gemini Image Generation and Editing},
  author       = {{Google}},
  year         = {2026},
  month        = oct,
  howpublished = {\url{https://deepmind.google/models/model-cards/nano-banana-2-1/}},
  note         = {Google Blog}
}

@misc{bytedance2026seedream5,
  title        = {Seedream 5.0: Deeper Thinking, More Accurate Generation},
  author       = {{ByteDance Seed}},
  year         = {2026},
  howpublished = {\url{https://seed.bytedance.com/en/blog/deeper-thinking-more-accurate-generation-introducing-seedream-5-0-lite}},
  note         = {ByteDance Seed Blog}
}

@misc{blackforestlabs2026flux3,
  title        = {FLUX 3: A Multimodal Frontier Model for Visual Intelligence},
  author       = {{Black Forest Labs}},
  year         = {2026},
  month        = jul,
  day          = {23},
  howpublished = {\url{https://bfl.ai/}},
  note         = {Official FLUX 3 model announcement}
}

@misc{google2025nanobanana,
  title        = {Introducing Gemini 2.5 Flash Image (aka nano-banana)},
  author       = {{Google}},
  year         = {2025},
  month        = aug,
  howpublished = {\url{https://developers.googleblog.com/introducing-gemini-2-5-flash-image/}},
  note         = {Google Blog}
}

@misc{google2025nanobananapro,
  title        = {Introducing Nano Banana Pro},
  author       = {{Google}},
  year         = {2025},
  month        = nov,
  howpublished = {\url{https://blog.google/technology/ai/nano-banana-pro/}},
  note         = {Google Blog}
}

@misc{seedream2025seedream4,
      title={Seedream 4.0: Toward Next-generation Multimodal Image Generation},
      author={Yunpeng Chen and Yu Gao and Lixue Gong and Meng Guo and Qiushan Guo and Zhiyao Guo and Xiaoxia Hou and Weilin Huang and Yixuan Huang and Xiaowen Jian and Huafeng Kuang and Zhichao Lai and Fanshi Li and Liang Li and Xiaochen Lian and Chao Liao and Liyang Liu and Wei Liu and Yanzuo Lu and Zhengxiong Luo and Tongtong Ou and Guang Shi and Yichun Shi and Shiqi Sun and Yu Tian and Zhi Tian and Peng Wang and Rui Wang and Xun Wang and Ye Wang and Guofeng Wu and Jie Wu and Wenxu Wu and Yonghui Wu and Xin Xia and Xuefeng Xiao and Shuang Xu and Xin Yan and Ceyuan Yang and Jianchao Yang and Zhonghua Zhai and Chenlin Zhang and Heng Zhang and Qi Zhang and Xinyu Zhang and Yuwei Zhang and Shijia Zhao and Wenliang Zhao and Wenjia Zhu},
      year={2025},
      eprint={2509.20427},
      archivePrefix={arXiv},
      primaryClass={cs.CV},
      url={https://arxiv.org/abs/2509.20427}
}

@misc{wang2025seededit3,
      title={SeedEdit 3.0: Fast and High-Quality Generative Image Editing},
      author={Peng Wang and Yichun Shi and Xiaochen Lian and Zhonghua Zhai and Xin Xia and Xuefeng Xiao and Weilin Huang and Jianchao Yang},
      year={2025},
      eprint={2506.05083},
      archivePrefix={arXiv},
      primaryClass={cs.CV},
      url={https://arxiv.org/abs/2506.05083}
}

@misc{blackforestlabs2025flux2,
      title={FLUX.2: Frontier Visual Intelligence},
      author={{Black Forest Labs}},
      year={2025},
      month={November},
      day={25},
      howpublished = {\url{https://bfl.ai/blog/flux-2}},
      note={Official FLUX.2 model announcement}
}

@misc{wu2025qwenimage,
      title={Qwen-Image Technical Report}, 
      author={Chenfei Wu and Jiahao Li and Jingren Zhou and Junyang Lin and Kaiyuan Gao and Kun Yan and Sheng-ming Yin and Shuai Bai and Xiao Xu and Yilei Chen and Yuxiang Chen and Zecheng Tang and Zekai Zhang and Zhengyi Wang and An Yang and Bowen Yu and Chen Cheng and Dayiheng Liu and Deqing Li and Hang Zhang and Hao Meng and Hu Wei and Jingyuan Ni and Kai Chen and Kuan Cao and Liang Peng and Lin Qu and Minggang Wu and Peng Wang and Shuting Yu and Tingkun Wen and Wensen Feng and Xiaoxiao Xu and Yi Wang and Yichang Zhang and Yongqiang Zhu and Yujia Wu and Yuxuan Cai and Zenan Liu},
      year={2025},
      eprint={2508.02324},
      archivePrefix={arXiv},
      primaryClass={cs.CV},
      url={https://arxiv.org/abs/2508.02324}, 
}

@misc{yin2025qwenimagelayered,
      title={Qwen-Image-Layered: Towards Inherent Editability via Layer Decomposition},
      author={Shengming Yin and Zekai Zhang and Zecheng Tang and Kaiyuan Gao and Xiao Xu and Kun Yan and Jiahao Li and Yilei Chen and Yuxiang Chen and Heung-Yeung Shum and Lionel M. Ni and Jingren Zhou and Junyang Lin and Chenfei Wu},
      year={2025},
      eprint={2512.15603},
      archivePrefix={arXiv},
      primaryClass={cs.CV},
      url={https://arxiv.org/abs/2512.15603}
}

@inproceedings{stablediffusion,
	title        = {High-resolution image synthesis with latent diffusion models},
	author       = {Rombach, Robin and Blattmann, Andreas and Lorenz, Dominik and Esser, Patrick and Ommer, Bj{\"o}rn},
	year         = 2022,
	booktitle    = {Proceedings of the IEEE/CVF conference on computer vision and pattern recognition},
	pages        = {10684--10695}
}

@article{podell2023sdxl,
	title        = {Sdxl: Improving latent diffusion models for high-resolution image synthesis},
	author       = {Podell, Dustin and English, Zion and Lacey, Kyle and Blattmann, Andreas and Dockhorn, Tim and M{\"u}ller, Jonas and Penna, Joe and Rombach, Robin},
	year         = 2023,
	journal      = {arXiv preprint arXiv:2307.01952}
}

@inproceedings{zheng2022image,
  title={Image inpainting with cascaded modulation gan and object-aware training},
  author={Zheng, Haitian and Lin, Zhe and Lu, Jingwan and Cohen, Scott and Shechtman, Eli and Barnes, Connelly and Zhang, Jianming and Xu, Ning and Amirghodsi, Sohrab and Luo, Jiebo},
  booktitle={European conference on computer vision},
  pages={277--296},
  year={2022},
  organization={Springer}
}

@article{zheng2025pixperfect,
  title={PixPerfect: Seamless Latent Diffusion Local Editing with Discriminative Pixel-Space Refinement},
  author={Zheng, Haitian and Yao, Yuan and Yu, Yongsheng and Zhou, Yuqian and Luo, Jiebo and Lin, Zhe},
  journal={arXiv preprint arXiv:2512.03247},
  year={2025}
}

@inproceedings{peebles2023scalable,
	title        = {Scalable diffusion models with transformers},
	author       = {Peebles, William and Xie, Saining},
	year         = 2023,
	booktitle    = {Proceedings of the IEEE/CVF International Conference on Computer Vision},
	pages        = {4195--4205}
}

@article{zhang2025uniser,
  title={UniSER: A Foundation Model for Unified Soft Effects Removal},
  author={Zhang, Jingdong and Zhang, Lingzhi and Liu, Qing and Chiu, Mang Tik and Barnes, Connelly and Wang, Yizhou and You, Haoran and Liu, Xiaoyang and Zhou, Yuqian and Lin, Zhe and Shechtman, Eli and Amirghodsi, Sohrab and Li, Xin and Wang, Wenping and Zhan, Xiaohang},
  journal={arXiv preprint arXiv:2511.14183},
  year={2025}
}

@inproceedings{chen2025unireal,
  title={Unireal: Universal image generation and editing via learning real-world dynamics},
  author={Chen, Xi and Zhang, Zhifei and Zhang, He and Zhou, Yuqian and Kim, Soo Ye and Liu, Qing and Li, Yijun and Zhang, Jianming and Zhao, Nanxuan and Wang, Yilin and Ding, Hui and Lin, Zhe and Zhao, Hengshuang},
  booktitle={Proceedings of the Computer Vision and Pattern Recognition Conference},
  pages={12501--12511},
  year={2025}
}

@article{kong2025token,
  title={Token Reduction Should Go Beyond Efficiency in Generative Models--From Vision, Language to Multimodality},
  author={Kong, Zhenglun and Li, Yize and Zeng, Fanhu and Xin, Lei and Messica, Shvat and Lin, Xue and Zhao, Pu and Kellis, Manolis and Tang, Hao and Zitnik, Marinka},
  journal={arXiv preprint arXiv:2505.18227},
  year={2025}
}

@inproceedings{bolya2023token,
	title        = {Token merging for fast stable diffusion},
	author       = {Bolya, Daniel and Hoffman, Judy},
	year         = 2023,
	booktitle    = {Proceedings of the IEEE/CVF conference on computer vision and pattern recognition},
	pages        = {4599--4603}
}

@inproceedings{wang2024attention,
	title        = {Attention-Driven Training-Free Efficiency Enhancement of Diffusion Models},
	author       = {Wang, Hongjie and Liu, Difan and Kang, Yan and Li, Yijun and Lin, Zhe and Jha, Niraj K and Liu, Yuchen},
	year         = 2024,
	booktitle    = {Proceedings of the IEEE/CVF Conference on Computer Vision and Pattern Recognition},
	pages        = {16080--16089}
}

@article{smith2024todo,
	title        = {ToDo: Token Downsampling for Efficient Generation of High-Resolution Images},
	author       = {Smith, Ethan and Saxena, Nayan and Saha, Aninda},
	year         = 2024,
	journal      = {arXiv preprint arXiv:2402.13573}
}

@article{chen2024deep,
  title={Deep compression autoencoder for efficient high-resolution diffusion models},
  author={Chen, Junyu and Cai, Han and Chen, Junsong and Xie, Enze and Yang, Shang and Tang, Haotian and Li, Muyang and Lu, Yao and Han, Song},
  journal={arXiv preprint arXiv:2410.10733},
  year={2024}
}

@inproceedings{nitzan2024lazy,
  title={Lazy diffusion transformer for interactive image editing},
  author={Nitzan, Yotam and Wu, Zongze and Zhang, Richard and Shechtman, Eli and Cohen-Or, Daniel and Park, Taesung and Gharbi, Micha{\"e}l},
  booktitle={European Conference on Computer Vision},
  pages={55--72},
  year={2024},
  organization={Springer}
}

@inproceedings{you2025layer,
  title={Layer-and Timestep-Adaptive Differentiable Token Compression Ratios for Efficient Diffusion Transformers},
  author={You, Haoran and Barnes, Connelly and Zhou, Yuqian and Kang, Yan and Du, Zhenbang and Zhou, Wei and Zhang, Lingzhi and Nitzan, Yotam and Liu, Xiaoyang and Lin, Zhe and Shechtman, Eli and Amirghodsi, Sohrab and Lin, Yingyan Celine},
  booktitle={Proceedings of the Computer Vision and Pattern Recognition Conference},
  pages={18072--18082},
  year={2025}
}

@article{kim2023bk,
	title        = {Bk-sdm: A lightweight, fast, and cheap version of stable diffusion},
	author       = {Kim, Bo-Kyeong and Song, Hyoung-Kyu and Castells, Thibault and Choi, Shinkook},
	year         = 2023,
	journal      = {arXiv preprint arXiv:2305.15798}
}

@inproceedings{fang2023structural,
	title        = {Structural pruning for diffusion models},
	author       = {Gongfan Fang and Xinyin Ma and Xinchao Wang},
	year         = 2023,
	booktitle    = {Advances in Neural Information Processing Systems}
}

@inproceedings{whalen2025early,
  title={Early-Bird Diffusion: Investigating and Leveraging Timestep-Aware Early-Bird Tickets in Diffusion Models for Efficient Training},
  author={Whalen, Lexington and Du, Zhenbang and You, Haoran and Li, Chaojian and Li, Sixu and Lin, Yingyan},
  booktitle={Proceedings of the Computer Vision and Pattern Recognition Conference},
  pages={7675--7684},
  year={2025}
}

@article{you2019drawing,
  title={Drawing early-bird tickets: Towards more efficient training of deep networks},
  author={You, Haoran and Li, Chaojian and Xu, Pengfei and Fu, Yonggan and Wang, Yue and Chen, Xiaohan and Baraniuk, Richard G and Wang, Zhangyang and Lin, Yingyan Celine},
  journal={arXiv preprint arXiv:1909.11957},
  year={2019}
}

@inproceedings{xu2018deepcache,
	title        = {Deepcache: Principled cache for mobile deep vision},
	author       = {Xu, Mengwei and Zhu, Mengze and Liu, Yunxin and Lin, Felix Xiaozhu and Liu, Xuanzhe},
	year         = 2018,
	booktitle    = {Proceedings of the 24th annual international conference on mobile computing and networking},
	pages        = {129--144}
}

@inproceedings{moura2019cache,
	title        = {Cache me if you can: Effects of DNS time-to-live},
	author       = {Moura, Giovane CM and Heidemann, John and Schmidt, Ricardo de O and Hardaker, Wes},
	year         = 2019,
	booktitle    = {Proceedings of the Internet Measurement Conference},
	pages        = {101--115}
}

@article{ma2024learning,
	title        = {Learning-to-Cache: Accelerating Diffusion Transformer via Layer Caching},
	author       = {Ma, Xinyin and Fang, Gongfan and Mi, Michael Bi and Wang, Xinchao},
	year         = 2024,
	journal      = {arXiv preprint arXiv:2406.01733}
}

@article{hwang2025tq,
  title={TQ-DiT: Efficient Time-Aware Quantization for Diffusion Transformers},
  author={Hwang, Younghye and Lee, Hyojin and Kang, Joonhyuk},
  journal={arXiv preprint arXiv:2502.04056},
  year={2025}
}

@inproceedings{chen2025q,
  title={Q-dit: Accurate post-training quantization for diffusion transformers},
  author={Chen, Lei and Meng, Yuan and Tang, Chen and Ma, Xinzhu and Jiang, Jingyan and Wang, Xin and Wang, Zhi and Zhu, Wenwu},
  booktitle={Proceedings of the Computer Vision and Pattern Recognition Conference},
  pages={28306--28315},
  year={2025}
}

@article{salimans2022progressive,
	title        = {Progressive distillation for fast sampling of diffusion models},
	author       = {Salimans, Tim and Ho, Jonathan},
	year         = 2022,
	journal      = {arXiv preprint arXiv:2202.00512}
}

@inproceedings{meng2023distillation,
	title        = {On distillation of guided diffusion models},
	author       = {Meng, Chenlin and Rombach, Robin and Gao, Ruiqi and Kingma, Diederik and Ermon, Stefano and Ho, Jonathan and Salimans, Tim},
	year         = 2023,
	booktitle    = {Proceedings of the IEEE/CVF Conference on Computer Vision and Pattern Recognition},
	pages        = {14297--14306}
}

@article{kang2024distilling,
	title        = {Distilling Diffusion Models into Conditional GANs},
	author       = {Kang, Minguk and Zhang, Richard and Barnes, Connelly and Paris, Sylvain and Kwak, Suha and Park, Jaesik and Shechtman, Eli and Zhu, Jun-Yan and Park, Taesung},
	year         = 2024,
	journal      = {ECCV 2024}
}

@inproceedings{sauer2025adversarial,
	title        = {Adversarial diffusion distillation},
	author       = {Sauer, Axel and Lorenz, Dominik and Blattmann, Andreas and Rombach, Robin},
	year         = 2025,
	booktitle    = {European Conference on Computer Vision},
	pages        = {87--103},
	organization = {Springer}
}

@misc{vantage2025aws_p548xlarge,
      title={p5.48xlarge Pricing and Specs – AWS EC2},
      author={{Vantage}},
      year={2025},
      howpublished = {\url{https://instances.vantage.sh/aws/ec2/p5.48xlarge?currency=USD}},
      note={Vantage Instances website; provides EC2 instance specs and pricing information}
}

@Misc{xFormers2022,
  author =       {Benjamin Lefaudeux and Francisco Massa and Diana Liskovich and Wenhan Xiong and Vittorio Caggiano and Sean Naren and Min Xu and Jieru Hu and Marta Tintore and Susan Zhang and Patrick Labatut and Daniel Haziza and Luca Wehrstedt and Jeremy Reizenstein and Grigory Sizov},
  title =        {xFormers: A modular and hackable Transformer modelling library},
  howpublished = {\url{https://github.com/facebookresearch/xformers}},
  year =         {2022}
}

@article{dao2023flashattention,
  title={Flashattention-2: Faster attention with better parallelism and work partitioning},
  author={Dao, Tri},
  journal={arXiv preprint arXiv:2307.08691},
  year={2023}
}

@manual{pytorch_scaled_dot_product_attention,
  title        = {torch.nn.functional.scaled\_dot\_product\_attention - PyTorch Documentation},
  organization = {PyTorch},
  url          = {https://docs.pytorch.org/docs/stable/generated/torch.nn.functional.scaled_dot_product_attention.html},
  year={2026}
}

@inproceedings{dmd,
  title={One-step diffusion with distribution matching distillation},
  author={Yin, Tianwei and Gharbi, Micha{\"e}l and Zhang, Richard and Shechtman, Eli and Durand, Fredo and Freeman, William T and Park, Taesung},
  booktitle={Proceedings of the IEEE/CVF conference on computer vision and pattern recognition},
  pages={6613--6623},
  year={2024}
}

@article{dmdv2,
  title={Improved distribution matching distillation for fast image synthesis},
  author={Yin, Tianwei and Gharbi, Micha{\"e}l and Park, Taesung and Zhang, Richard and Shechtman, Eli and Durand, Fredo and Freeman, Bill},
  journal={Advances in neural information processing systems},
  volume={37},
  pages={47455--47487},
  year={2024}
}

@article{deng2020model,
	title        = {Model compression and hardware acceleration for neural networks: A comprehensive survey},
	author       = {Deng, Lei and Li, Guoqi and Han, Song and Shi, Luping and Xie, Yuan},
	year         = 2020,
	journal      = {Proceedings of the IEEE},
	publisher    = {IEEE},
	volume       = 108,
	number       = 4,
	pages        = {485--532}
}

@article{wang2020dual,
	title        = {Dual dynamic inference: Enabling more efficient, adaptive, and controllable deep inference},
	author       = {Wang, Yue and Shen, Jianghao and Hu, Ting-Kuei and Xu, Pengfei and Nguyen, Tan and Baraniuk, Richard and Wang, Zhangyang and Lin, Yingyan},
	year         = 2020,
	journal      = {IEEE Journal of Selected Topics in Signal Processing},
	publisher    = {IEEE},
	volume       = 14,
	number       = 4,
	pages        = {623--633}
}

@inproceedings{wu2018blockdrop,
	title        = {Blockdrop: Dynamic inference paths in residual networks},
	author       = {Wu, Zuxuan and Nagarajan, Tushar and Kumar, Abhishek and Rennie, Steven and Davis, Larry S and Grauman, Kristen and Feris, Rogerio},
	year         = 2018,
	booktitle    = {Proceedings of the IEEE conference on computer vision and pattern recognition},
	pages        = {8817--8826}
}

@inproceedings{wang2018skipnet,
	title        = {Skipnet: Learning dynamic routing in convolutional networks},
	author       = {Wang, Xin and Yu, Fisher and Dou, Zi-Yi and Darrell, Trevor and Gonzalez, Joseph E},
	year         = 2018,
	booktitle    = {Proceedings of the European conference on computer vision (ECCV)},
	pages        = {409--424}
}

@article{zhao2024dynamic,
	title        = {Dynamic diffusion transformer},
	author       = {Zhao, Wangbo and Han, Yizeng and Tang, Jiasheng and Wang, Kai and Song, Yibing and Huang, Gao and Wang, Fan and You, Yang},
	year         = 2024,
	journal      = {arXiv preprint arXiv:2410.03456}
}

@article{raposo2024mixture,
	title        = {Mixture-of-Depths: Dynamically allocating compute in transformer-based language models},
	author       = {Raposo, David and Ritter, Sam and Richards, Blake and Lillicrap, Timothy and Humphreys, Peter Conway and Santoro, Adam},
	year         = 2024,
	journal      = {arXiv preprint arXiv:2404.02258}
}

@inproceedings{teerapittayanon2016branchynet,
	title        = {Branchynet: Fast inference via early exiting from deep neural networks},
	author       = {Teerapittayanon, Surat and McDanel, Bradley and Kung, Hsiang-Tsung},
	year         = 2016,
	booktitle    = {2016 23rd international conference on pattern recognition (ICPR)},
	pages        = {2464--2469},
	organization = {IEEE}
}

@article{huang2017multi,
	title        = {Multi-scale dense networks for resource efficient image classification},
	author       = {Huang, Gao and Chen, Danlu and Li, Tianhong and Wu, Felix and Van Der Maaten, Laurens and Weinberger, Kilian Q},
	year         = 2017,
	journal      = {arXiv preprint arXiv:1703.09844}
}

@inproceedings{mullapudi2018hydranets,
	title        = {Hydranets: Specialized dynamic architectures for efficient inference},
	author       = {Mullapudi, Ravi Teja and Mark, William R and Shazeer, Noam and Fatahalian, Kayvon},
	year         = 2018,
	booktitle    = {Proceedings of the IEEE Conference on Computer Vision and Pattern Recognition},
	pages        = {8080--8089}
}

@inproceedings{moon2023early,
	title        = {Early exiting for accelerated inference in diffusion models},
	author       = {Moon, Taehong and Choi, Moonseok and Yun, EungGu and Yoon, Jongmin and Lee, Gayoung and Lee, Juho},
	year         = 2023,
	booktitle    = {ICML 2023 Workshop on Structured Probabilistic Inference $\&$ Generative Modeling}
}

@article{ye2025imgedit,
  title={Imgedit: A unified image editing dataset and benchmark},
  author={Ye, Yang and He, Xianyi and Li, Zongjian and Lin, Bin and Yuan, Shenghai and Yan, Zhiyuan and Hou, Bohan and Yuan, Li},
  journal={arXiv preprint arXiv:2505.20275},
  year={2025}
}

@article{labs2025flux,
  title={FLUX. 1 Kontext: Flow Matching for In-Context Image Generation and Editing in Latent Space},
  author={{Black Forest Labs} and Batifol, Stephen and Blattmann, Andreas and Boesel, Frederic and Consul, Saksham and Diagne, Cyril and Dockhorn, Tim and English, Jack and English, Zion and Esser, Patrick and Kulal, Sumith and Lacey, Kyle and Levi, Yam and Li, Cheng and Lorenz, Dominik and M{\"u}ller, Jonas and Podell, Dustin and Rombach, Robin and Saini, Harry and Sauer, Axel and Smith, Luke},
  journal={arXiv preprint arXiv:2506.15742},
  year={2025}
}
\bibliographystyle{mlsys2025}

% \appendix
% \section{Appendix}
% \label{sec:appendix}

\end{document}